\documentclass[sigconf, review=False]{acmart}

\usepackage{booktabs} % For formal tables
\usepackage{subcaption}
\usepackage{multirow}
\setcopyright{rightsretained}
\usepackage{booktabs,caption}
\usepackage[flushleft]{threeparttable}
\usepackage{balance} 

\acmDOI{10.1145/3219819.3219930}

\acmISBN{978-1-4503-5552-0/18/08}

\acmConference[KDD 2018]{Knowledge Discovery and Data Mining}{August 2018}{London, UK}
\acmYear{2018}
\copyrightyear{2018}

\acmArticle{4}
\acmPrice{15.00}

\begin{document}

\title{Learning Tasks for Multitask Learning: Heterogenous Patient Populations in the ICU}

\author{Harini Suresh} \authornote{\label{author}The first two authors contributed equally to this work.}
\affiliation{%
	\institution{Massachusetts Institute of Technology}
    \department{Computer Science and Artificial Intelligence Laboratory}
    %\city{Cambridge}
    %\state{MA}
    %\postcode{02139}
    %\country{USA}
}
\email{hsuresh@mit.edu}

\author{Jen J. Gong}\authornotemark[1]
\affiliation{%
	\institution{Massachusetts Institute of Technology}
    \department{Computer Science and Artificial Intelligence Laboratory}
    %\city{Cambridge}
    %\state{MA}
    %\postcode{02139}
    %\country{USA}
}
\email{jengong@mit.edu}

\author{John V. Guttag}
\affiliation{%
	\institution{Massachusetts Institute of Technology}
    \department{Computer Science and Artificial Intelligence Laboratory}
    %\city{Cambridge}
    %\state{MA}
   % \postcode{02139}
    %\country{USA}
}
\email{guttag@mit.edu}

\begin{abstract}
 
Machine learning approaches have been effective in predicting adverse outcomes in different clinical settings. These models are often developed and evaluated on datasets with heterogeneous patient populations. However, good predictive performance on the aggregate population does not imply good performance for specific groups.

In this work, we present a two-step framework to 1) learn relevant patient subgroups, and 2) predict an outcome for separate patient populations in a multi-task framework, where each population is a separate task. We demonstrate how to discover relevant groups in an unsupervised way with a sequence-to-sequence autoencoder. We show that using these groups in a multi-task framework leads to better predictive performance of in-hospital mortality both across groups and overall. We also highlight the need for more granular evaluation of performance when dealing with heterogeneous populations.

\end{abstract}




\maketitle

\section{Introduction}

Many important applications of machine learning utilize data from groups with different characteristics. Models trained on these datasets may not result in good predictions for each constituent group. This has been illustrated in tasks such as image classification \cite{46553}, face recognition \cite{pmlr-v81-buolamwini18a}, and advertising \cite{datta2015automated}. In this work, we investigate this problem in clinical data, where such datasets are prevalent. 

Machine learning models developed for clinical prediction tasks have the ability to aid care staff in deciding appropriate treatments. However, these clinical decision-making tools typically are not developed with specific subpopulations in mind, or they are developed for a single subpopulation and can suffer from data scarcity. The existence of these different subpopulations gives rise to a multifaceted problem: 1) a single model built for the entire patient population in aggregate does not imply equally good performance across distinct patient subpopulations, and 2) separate models learned on each of the distinct patient subpopulations do not take advantage of the shared knowledge that is common across patient subgroups.

Our solution combines \textit{cohort discovery} with a \textit{multi-task learning} model. Together, these steps form a pipeline that leverages shared information across distinct patient cohorts while accounting for their differences. During cohort discovery, we learn distinct patient subgroups in a data-driven way. These groups allow us to utilize a multi-task prediction framework where distinct patient groups are separate {\it tasks}. In order for multi-task learning to work effectively in this setup, examples need to be grouped into subpopulations that are sufficiently distinct with relation to the outcome of interest so that separate task models are beneficial.

Task formulations for multi-task learning with clinical data fall into two categories: 1) distinct {\it outcomes} are used as tasks~\cite{Caruana1993MultitaskLA, harutyunyan2017multitask, razavian2016multi, wiens2016patient} and 2) distinct {\it patient populations} are used as tasks \cite{nori2017learning, xu2015formula}. Our formulation falls in the second category, where different {\it patient populations} are regarded as different tasks. Prior work has investigated pre-defined task definitions (e.g., \cite{wiens2016patient}), and other work has used billing diagnosis codes to define latent bases for each patient \cite{nori2017learning}. In this work, we use physiological time-series dynamics to group examples into meaningful clinical tasks.

We investigate these methods in the context of building predictive models for patients in intensive care units (ICUs), using data from the publicly available MIMIC-III intensive care dataset~\cite{johnson2016mimic}.

Although patients in the ICU are typically more severely ill than patients in the hospital at large, the heterogeneity of patients in the ICU provides a useful case study for our approach, and MIMIC, as a publicly accessible dataset, enables reproducible studies. 

We focus on the task of predicting whether a patient will die in the hospital, using data from the initial duration (24 hours or 48 hours) of their stay. Mortality prediction is an important task in clinical settings because a high risk of mortality is a signal of declining state and need for intervention. We show that a) there are salient subpopulations in the data that we can discover, and b) a multi-task model with subpopulations as tasks can outperform a single model that ignores subpopulation differences (a {\it global} model) as well as a single model trained on each subpopulation ({\it separate} models) on both overall and per-group performance metrics. 

We also demonstrate the importance of performing granular evaluations across important subpopulations in a dataset. While much work reports overall metrics of performance, we highlight how this can hide underperformance on specific groups.

In Section~\ref{related_work}, we describe the literature in machine learning for healthcare pertaining to 1) patient cohort discovery, and 2) multi-task learning. In Section~\ref{data}, we describe the data we use. Next, we describe our two-step model formulation in Section~\ref{methods}, and our experiments and results in Sections~\ref{experiments} and ~\ref{results}.

\section{Related Work}
\label{related_work}

% \subsection{Clinical Data Analysis}
% The rapid adoption and availability of electronic health records (EHRs) has given way to new investigations into data-driven clinical support [citation]. The broad goal of these studies is to learn from massive datasets of patient records in order provide personalized treatment to current and future patients. Studies have attempted to predict clinical events from various interventions [cite], to diagnoses [cite], to mortality [cite], and have used data including physiological variables [cite], lab test results [cite], unstructured clinical notes [cite], and procedures [cite]. 

The rapid adoption and availability of electronic health records (EHRs) has enabled new investigations into data-driven clinical support \cite{wu2010prediction, obermeyer2016predicting, ghassemi2015state}. The broad goal of these studies is to learn from datasets of patient records in order to provide personalized treatment to patients. We provide a brief overview of work specifically in patient cohort discovery and multi-task learning. 

\subsection{Patient Cohort Discovery}

Work in patient cohort discovery has focused on finding phenotypic characteristics of patients relevant for clinical insights, diagnoses, or risk-stratification. Constructing these groups requires finding a robust and meaningful representation of a patient's state. 

\subsubsection{Patient Representations}
Static risk scores such as the Simplified Acute Physiology Score (SAPS II) \cite{le1993new} can be used to characterize a patient's state; these scores use a limited number of variables and do not take into account temporal trends \cite{sinuff2006mortality}.  Many recent works aim to \textit{learn} data-driven representations of a patient's state. Some of these are learned in a supervised framework: for example, using the representation learned in a hidden layer of a deep neural network as a representation of patient state \cite{che2015deep}.  Other works characterize evolving patient state in an unsupervised way, inferring topics from clinical notes using Latent Dirichlet Allocation (LDA) \cite{ghassemi2014unfolding}, or inferring states and transitions with a switching state autoregressive model \cite{ghassemi2017predicting}.
%%%%%%%%%%%%%
\begin{figure*}[t!]
\includegraphics[scale=0.6]{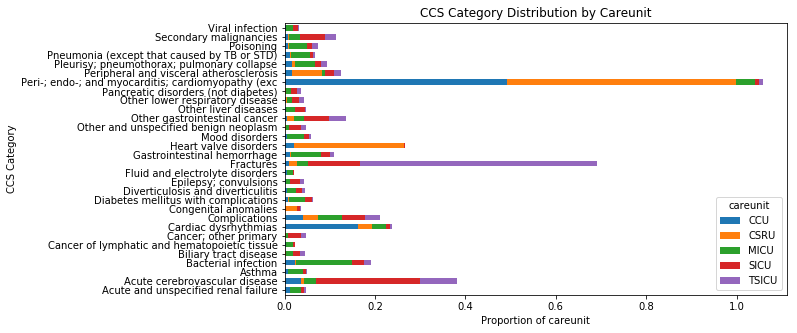}
\caption{Primary diagnoses for patient admissions by Clinical Classifications Software (CCS) categories.}
\label{ccs_categories}
\end{figure*}
%%%%%%%%%%%%%%

\subsubsection{Cohort Discovery}\label{cohort_discovery}
After constructing a meaningful representation, \textit{cohort discovery} requires using this representation to group patients into relevant cohorts. There is a broad range of what is considered a cohort (sometimes referred to as a \textit{phenotype} in the literature) and how they are learned. In some cases, cohorts are pre-defined: for example, \citeauthor{gehrmann2017comparing} have a group of physicians manually annotate examples with a set of 10 disease-related cohort classifications \cite{gehrmann2017comparing}. The process of manual annotation, however, is time-consuming, expensive and hard to scale. With the growing availability of large, high-dimensional clinical data, many works have proposed approaches to learning patient phenotypes~\cite{ho2014limestone,ho2014marble,pivovarov2015learning,che2015deep}. 
In all of these works, the patient cohorts are either analyzed for clinical insight, or used as additional features in a supervised prediction problem with a single, global model. In contrast to these works, we use the learned cohorts in a multi-task framework so that we can explicitly optimize for performance on each cohort.

\subsection{Multi-task Learning for Clinical Risk-Stratification}
The goal of multi-task learning is to combine learning of multiple related tasks, in order to improve performance across tasks (as opposed to learning each independently). \citeauthor{zhang2017survey} present a comprehensive overview of multi-task methods \cite{zhang2017survey}, and \citeauthor{DBLP:journals/corr/Ruder17a} give an overview of implementations of multi-task learning with deep neural networks \cite{DBLP:journals/corr/Ruder17a}. 

In the clinical space, multi-task models have been used in a framework where the tasks are different prediction problems: for example,  \citeauthor{harutyunyan2017multitask} train a multi-task recurrent neural network that predicts mortality, length of stay, and ICD-9 groupings \cite{harutyunyan2017multitask}, \citeauthor{razavian2016multi} compare multi-task convolution and recurrent neural networks for predicting a number of ICD-9 diagnoses \cite{razavian2016multi}, and \citeauthor{choi2016doctor} use recurrent neural network architecture to predict both diagnoses and the duration until the next visit \cite{choi2016doctor}. \citeauthor{ngufor2015heterogeneous} use a multi-task model to improve prediction of various outcomes related to surgical procedures \cite{ngufor2015heterogeneous}.
\citeauthor{wang2014exploring} directly compare a multi-task model with many single-task models to demonstrate the utility of transferring knowledge across tasks for disease prediction \cite{wang2014exploring}. Other work has explored post-learning strategies to cluster similar tasks in a multi-task model to enable optimal cross-transfer of knowledge \cite{7344836}. Hierarchical models have also been used to predict multiple outcomes \cite{shaddox2016hierarchical}.

Predicting multiple outcomes aims to improve the generalizability of a model, whereas our goal is to build the best-suited model for distinct patient subpopulations by using the populations as the different tasks. \citeauthor{nori2017learning} do this by constructing a small number of latent basis tasks each with their own parameter vectors, and representing each patient as a combination of these tasks \cite{nori2017learning}. The specific combination is determined by the patient's record of diseases, represented as ICD-10 codes. Similarly, \cite{xu2015formula} uses a framework where patient-specific tasks are formulated as a linear combination of a shared set of base models. We consider salient and characterizable patient subpopulations, rather than separate tasks for each individual patient.

Other work aims to identify patient cohorts and transfer knowledge between them in a prediction framework. For example, hierarchical models have been used to take into account population differences \cite{corbin2012hierarchical,d2007use,alaa2016personalized}. \citeauthor{alaa2016personalized} discover latent ``classes'' in the data, train Gaussian Processes to model the physiological data stream for each class, and transfer knowledge learned about the clinically stable population to a clinically declining population \cite{alaa2016personalized}. Our method has a similar aim (discovering groups in the data and utilizing shared knowledge across these groups) though we do not assume the framework of transferring knowledge from clinically stable to declining populations.

% \citeauthor{NIPS2011_4292} use clustered multi-task learning (CMTL) to simultaneously cluster tasks and make predictions, in order to share more information across related tasks \cite{NIPS2011_4292}. On the other had, we aim to cluster individual \textit{examples} into tasks, allowing us to learn distinct tasks to use in an multi-task model.

Our two-step pipeline enables us to learn patient subgroups that we use as tasks in a multi-task framework. In addition, it leverages the underlying physiological data of the patient rather than domain knowledge or auxiliary labels to discover relevant patient cohorts.

\section{Data}
\label{data}

\begin{table}[b!]
\caption{Number of adult patients and rate of in-hospital mortality (defined using the earliest time of mortality, or a note of ``do not resuscitate" (DNR) or ``comfort-measures only" (CMO) in each intensive care unit (ICU).}
\resizebox{0.5\textwidth}{!}{
\begin{tabular}{l |rrrrrr}
\toprule
{\bf Careunit} & $N$ & $n$  & Class Imbalance & Age (Mean) & Gender (Male)\\
\midrule
{\bf CCU    } &       4,905 &  339 &  0.069 & 82.56 &   0.58  \\
{\bf CSRU   }  &       6,969 & 137 &  0.020 & 69.46 &   0.67 \\
{\bf MICU   } &      11,395 &  1118 &  0.098 & 77.98 &   0.51  \\
{\bf SICU   }  &       5,178 &  397 & 0.077& 72.57&   0.52  \\
{\bf TSICU }  &       4,239 &  283 &  0.067  & 67.14 &   0.61\\ \hline
{\bf Overall}  &      32,686 & 2274  & 0.070 & 74.59 &   0.57  \\
\bottomrule
\end{tabular}
\label{N_n}
}
\end{table}

\begin{table}[b!]
\centering
\caption{Physiological variables used for prediction.}
\resizebox{0.5\textwidth}{!}{
\begin{tabular}{c|ccc}
\hline
Static Variables & Gender & Age & Ethnicity \\
\hline
Vitals and Labs & Anion gap & Bicarbonate & blood pH\\
 & Blood urea nitrogen & Chloride & Creatinine\\
 & Diastolic blood pressure & Fraction inspired oxygen & Glascow coma scale total\\
 & Glucose & Heart rate & Hematocrit\\
 & Hemoglobin & INR\textsuperscript{*} & Lactate\\
 & Magnesium & Mean blood pressure & Oxygen saturation\\
 & Partial thromboplastin time & Phosphate & Platelets\\
 & Potassium & Prothrombin time & Respiratory rate\\
 & Sodium & Systolic blood pressure & Temperature\\
 & Weight & White blood cell count\\
\hline
\multicolumn{4}{l}{\textsuperscript{*}\footnotesize{International normalized ratio of the prothrombin time}}
\end{tabular}
}
\label{tab:variables}
\end{table}

We use data from the publicly available MIMIC-III database \cite{johnson2016mimic}. Although MIMIC-III primarily contains data from a critical care setting, it has a large, heterogeneous patient population, and the conclusions we draw from it in this work are likely relevant considerations for prediction tasks in other clinical settings. In addition, the dataset is made publicly available to researchers, enabling reproducibility. The dataset contains both structured electronic health record-like data, as well as free text clinical notes. We utilize the highly sampled vitals signs and irregularly sampled lab test results from the structured data, as well as static demographic attributes such as age, gender, and ethnicity. Table \ref{tab:variables} contains a full list of features used in our experiments. Prior work has used these time-series to understanding patient physiological state to predict various outcomes such as intervention administration and mortality~\cite{wu2017understanding,ghassemi2017predicting, suresh2017clinical,che2015deep}. 

Patient characteristics are summarized in Table~\ref{N_n} and Figure~\ref{ccs_categories}. In particular, we note that the patients in different care units have very different rates of mortality, ranging from 2.0\% in the Cardiac Surgery Recovery Unit (CSRU) to 9.8\% in the Medical Intensive Care Unit (MICU). In addition, we note that patients in different units often present with different conditions, from acute events such as bone fractures to chronic conditions such as hypertension and coronary artery disease. Figure~\ref{ccs_categories} shows the presence of some different disease categories.

\section{Methods}
\label{methods}
\begin{figure*}[t!]
\includegraphics[width=.9\textwidth]{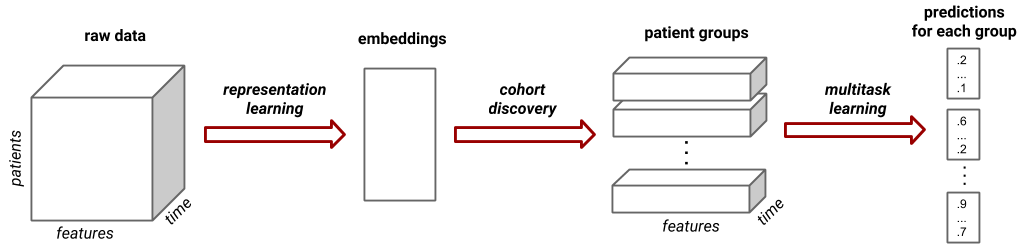}
\caption{We present a two-step pipeline for 1) discovering relevant cohorts from the underlying physiological data for the prediction task at hand, and 2) using multi-task learning to share knowledge across related data while allowing distinct models to make predictions for different patient populations.}
\label{fig:flowchart}
\end{figure*}

\begin{figure*}
	\centering
    \begin{subfigure}[t]{0.33\textwidth}
    \centering\captionsetup{width=.8\linewidth}
        \includegraphics[width=\columnwidth]{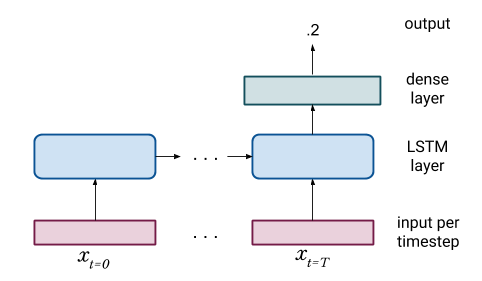}
        \caption[width=.2\textwidth]{Single task model that does not differentiate between groups.}\label{fig:single_task_model}
    \end{subfigure}%
 ~
    \begin{subfigure}[t]{0.33\textwidth}
        \centering\captionsetup{width=.8\linewidth}
        \includegraphics[width=\columnwidth]{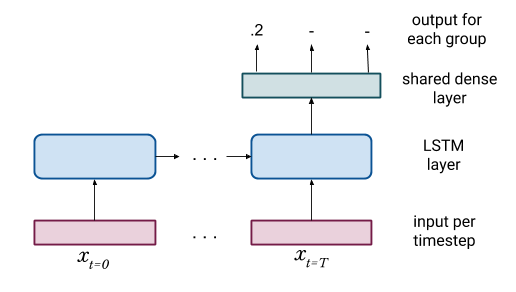}
        \caption{Multi-task model with separate parameters for each group at the final output layer.}\label{fig:mtl_shared_dense}
    \end{subfigure}%
        ~
            \begin{subfigure}[t]{0.33\textwidth}
        \centering\captionsetup{width=.8\linewidth}
        \includegraphics[width=\columnwidth]{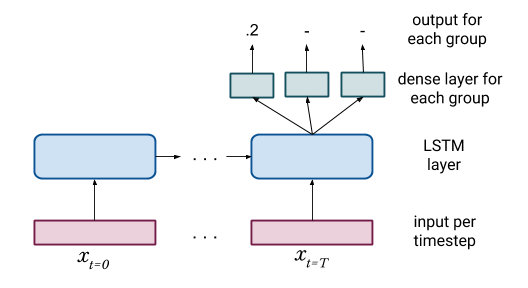}
        \caption{Multi-task model with separate dense layers for each group.} \label{fig:mtl_sep_dense}
    \end{subfigure}

\caption{Single task and multi-task model configurations. Single task models have shared parameters for all examples, while multi-task models have separate parameters for each group in the output layer and/or the final dense layer.}
\label{fig:models}
\end{figure*}

In this section, we describe our two-step procedure for 1) identifying meaningful patient cohorts, and 2) leveraging these cohorts as separate tasks in a multi-task learning framework.~\footnote{Model code is available at github.com/mit-ddig/multitask-patients.} This pipeline is diagrammed in Figure~\ref{fig:flowchart}.

\subsection{Identifying Meaningful Patient Cohorts}

We utilize unsupervised representations and cohort-discovery methods for identifying relevant patient cohorts. Importantly, this method relies only on attributes at hospital admission or data from the initial portion of the patient's stay. Using this interval of data for patient phenotyping allows us to 1) identify patient phenotypes that are relevant when longitudinal patient history may not be immediately available, and 2) utilize only information prior to the time at which we make a prediction.

The raw patient data is a sparse timeseries; in order to discover cohorts we first encode this raw data into a dense fixed-length representation that we then cluster. We use a long short-term memory (LSTM) \citep{hochreiter1997long} autoencoder to produce a dense representation that captures important facets of the input. LSTMs have effectively modeled complicated dependencies in many types of time-series data \cite{bengio1994learning, chorowski2015attention, hermann2015teaching, xu2015show}, including clinical time-series \citep{suresh2017clinical, che2016recurrent,lipton2015learning,razavian2016multi}. They are well-suited to our task because of the complex temporal dependencies in physiological time-series. Autoencoders have been used to learn compact representations of patient state from multi-modal timeseries EHR data \cite{suresh2017use,miotto2016deep}.

We use one LSTM layer in the encoder and one in the decoder. The embedded representation is the state of the encoder LSTM at the final timestep. This representation is then used in the decoder to reconstruct the input timeseries. 

Embedding size was tuned for the reconstruction loss based on the training and validation data. Because reconstruction loss will consistently decrease when the embedding size is increased, we chose the embedding size based on the elbow in the reconstruction curve. We then cluster the embeddings with a  Gaussian Mixture Model (GMM). The cluster assignments are used to group patients into tasks for the multi-task model.

\subsection{Learning Predictive Models}

In the prediction step, in order to go from a patient timeseries to a mortality prediction, we use an LSTM for all of the model configurations.

Our proposed approach uses a {\it multi-task} model, and we compare against several {\it single-task} baselines. The differences in these model configurations and training procedures are discussed in this section.

\subsubsection{Single Task Model}
The single task model (Figure \ref{fig:single_task_model}) consists of a single LSTM layer with a ReLU activation function followed by a single fully-connected layer with a sigmoid activation function. The output of the fully-connected layer is an estimate of the probability of mortality for the given example. We train this single task model on all the data to produce the global model baseline, and separately on data from each group to produce the separate model baselines.

\subsubsection{Multi-task Model}

In the multi-task model, our goal is to combine shared, global parameters along with separate parameters trained specifically for each group. In order to do this, we use the hard parameter-sharing framework of multi-task learning introduced in \cite{Caruana1993MultitaskLA}.

Like the single-task model, the multi-task model has one LSTM layer. The multi-task model was used either a single separate fully-connected layer for each group (Figure~\ref{fig:mtl_sep_dense}) {\it or} a shared dense layer with separate weights leading to the output ((Figure \ref{fig:mtl_shared_dense}). During our grid search for model configurations, we limited the size of these fully-connected layers compared to the fully-connected layer of the single task model to ensure that both configurations were able to have similar capacity for making a fair comparison. The task-specific parameters are trained using only the losses from examples belonging to the task. 

We compared our multi-task learning approach against two single-task approaches: 1) a separate single task model for each group, and 2) a global model for all patients, agnostic to task membership.

\subsection{Evaluating Predictive Models Across Patient Cohorts}
Machine learning models for clinical outcome predictions often utilize aggregate discriminative metrics such as the area under the receiver operating characteristic curve (AUC) to account for class imbalance (e.g., \cite{che2015deep, suresh2017clinical,gong2017predicting}). In settings where evaluations on specific patient cohorts is of interest, evaluation is more challenging. To evaluate metrics over different populations or outcomes, {\it micro} and {\it macro} versions of predictive metrics are used. In the {\it micro} case, all of the predicted probabilities for all patients are treated as if they come from a single classifier:

\begin{equation}
Metric_{micro} = Metric \big( [\hat{y}_0, \cdots, \hat{y}_K],  [y_0, \cdots, y_K] \big),\\
\end{equation}

\noindent where $K$ = the number of groups, $\hat{y}_k$ = predictions for the examples in group $k$ and $y_k$ = true labels for the examples in group $k$. This is the metric that is typically used in the literature. However, using these micro-evaluated metrics makes it difficult to assess how a model is performing on different subpopulations. This is especially true when the subpopulations are not equally represented.

{\it Macro} measures evaluate a metric {\it within} each cohort first, and then average the results across cohorts: 

\begin{equation}
Metric_{macro} = \frac{1}{K}\sum_{k=0}^K Metric \big( \hat{y}_k, y_k \big)\\
\end{equation}

This metric is better suited to assess performance across groups of disparate size, since each group contributes equally to the macro metric evaluation \cite{macro_auc}.  

We use both of these methods of computing metrics, and evaluate micro- and macro- AUC. We additionally evaluate micro- and macro- positive predictive value (PPV) and specificity at a sensitivity of 80\%. While AUC gives a sense of overall discriminative model performance, we show PPV and specificity at a single decision threshold to evaluate how well such a model might perform in a real setting.

\section{Experiments}
\label{experiments}

We developed models for predicting in-hospital mortality using physiological time-series data from the initial portion of the patient's ICU stay. 

\subsection{Prediction Task Definition}
We define {\it in-hospital mortality} as having an outcome of mortality, {\it or} a note of "Do Not Resuscitate" (DNR) or "Comfort Measures Only" (CMO). This definition is in contrast to what has been used in prior work, where only mortality was considered (e.g., ~\cite{ghassemi2014unfolding,harutyunyan2017multitask,gong2017predicting}). Notes of DNR or CMO indicate differences in what clinical interventions will be taken, and our proposed risk models might not have actionable predictions.

We conducted experiments in two settings: 
\begin{enumerate}
\item Using the first 24 hours of data from the patient's stay to predict in-hospital mortality starting at 36 hours into the stay. Acuity scores such as the Simplified Acute Physiology Score (SAPS-II)~\cite{le1993new} also use the first 24 hours of data to evaluate patient severity of illness.
\item Using the first 48 hours of data from the patient's stay to predict in-hospital mortality starting 72 hours in to the stay. We explore this task because the first 24 hours often contain routine tests done upon admission, and this time period might reflect different changes in patient physiology. 
\end{enumerate}

Each of the described experiments includes prediction gaps between the information used about a patient and the point at which outcomes are counted. This is common in the literature, and the motivation is two-fold: 1) it eliminates trivial cases where the outcome is imminent, and 2) simulates a situation in which there is time to intervene. Patients who were discharged or had an outcome of in-hospital mortality during the period of the stay being used for prediction or during the gap period were dropped from the experiments.

\subsection{Data Processing}

We considered all ICU patients over the age of 15 and took the patient's first ICU stay (if there are multiple), as the majority of patients have only one ICU stay.  For each patient, we extracted 29 time-varying vitals and labs, detailed in Table~\ref{tab:variables}. The timestamps of these measurements were rounded to the nearest hour. If an hour had multiple measurements for a signal, those measurements were averaged. We created discrete, binary features by first transforming each variable to the $z$-score, and then making each $z$-score value its own column. Similar methods have been used in previous work \cite{suresh2017use,wu2017understanding} in order to have an explicit representation of missing values, and to avoid overfitting to small changes in the physiological variables. This creates a very sparse data representation. 

In addition, we include demographics such as the patient's ethnicity, gender and age quartile. These static variables are replicated across all time-steps for a patient.

\begin{figure*}[t!]
\includegraphics[scale=0.4]{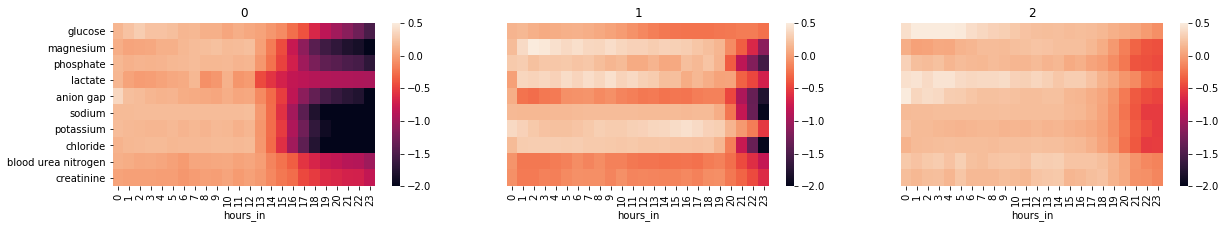}
\includegraphics[scale=0.4]{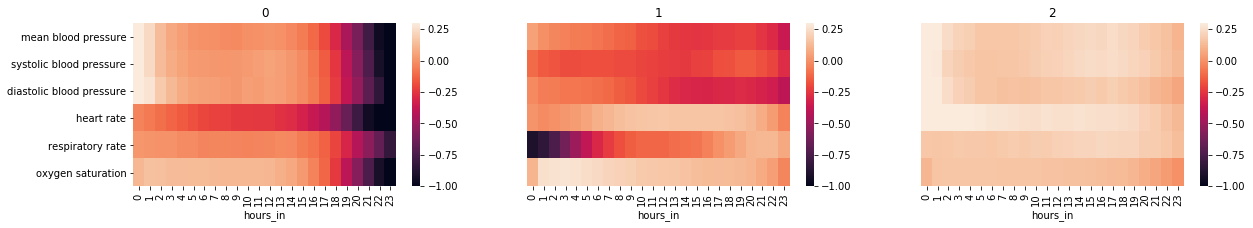}
\caption{Selected lab test and vital signs features over the first 24 hours for the unsupervised clusters. Figures show $z$-score values of the features over time; 0 indicates the mean value, positive values indicate elevated measures, and negative values indicate decreased measures. In the first 24 hours, lab test results in cluster 2 are more elevated than in cluster 0 and cluster 1. Cluster 0 has a centroid with decreasing heart rate, whereas cluster 1's centroid shows an increasing heart rate. Note that while the trends are opposing, both centroids have heart rate values that are below the mean. Additionally, we note that blood test results for magnesium, lactate, and potassium are elevated in cluster 1, while glucose is elevated in cluster 2.}
\label{phys_24_48}
\end{figure*}

\subsection{Model Implementation and Training}

In this section, we describe our model training and selection procedures. We describe 1) the supervised models trained for predicting the outcome using a single-task and multi-task framework, 2) the autoencoders used to learn unsupervised, latent representations for our physiological data, and 3) the Gaussian Mixture Models used to identify cohorts from the latent representations of the data. For all experiments, we split the data using an 80:20 training:test split stratified on the outcome. 

\subsubsection{Gaussian Mixture Model} We used the Scikit-learn (version 0.19.1) implementation of Gaussian Mixture Models \cite{scikit-learn}. The GMM was initialized using assignments from k-means clustering. We fit the model with 30 different initializations and chose the model that gave the highest data likelihood. We divided the training data in a 7:1 training:validation split. We explored several possible values for the number of clusters, and chose the value that resulted in the best predictive performance on the validation set when the clusters were used as tasks in the multi-task model.

\subsubsection{Unsupervised Representations} To learn unsupervised representations, we used a sequence-to-sequence autoencoder with LSTM units implemented with Keras.  We explored several hidden dimension sizes for the autoencoder, and chose the dimensionality corresponding to the elbow in the reconstruction error curve on the validation set. This procedure resulted in an embedding size of 100. 

The autoencoder was trained with a mean squared error loss function and the Adam optimizer with an initial learning rate of 0.001. We trained the autoencoder for a maximum of 100 epochs; to prevent over-training, we employed early stopping if the validation loss decreased for 6 epochs.

\subsubsection{Single and Multi-task Prediction Models}\label{model-setup} We implemented the single and multi-task models using Keras version 2.1.3 \cite{chollet2015keras}. We determined the best model configurations by doing a grid search over possible hyperparameters and choosing the best configuration over 5 random splits of the training data into 7:1 training:validation splits. We allowed the global model to search over a larger range of layer sizes to enable a fair comparison with the extra parameters that could be introduced in the multi-task model. We used binary cross-entropy as our loss function, and the Adam optimizer with a learning rate of 0.0001. The models were trained for a maximum of 100 epochs with early stopping.

\section{Results}
\label{results}

We report results comparing the global single-task model with the multi-task model. We also tested a baseline of using separate single-task models for each task, but this model had significantly worse performance in all cases so we have not included it. All reported statistical significance results were computed using the Wilcoxon signed-rank test~\cite{wilcoxon1945individual} over 100 bootstrapped samples of the test set. Bootstrapped samples were of the same size and class imbalance as the original test set.

\begin{table}
\caption{Cohort statistics at 24 hours and 48 hours}
 \resizebox{0.5\textwidth}{!}{
 \begin{centering}
\begin{tabular}{c |c| c|c|c | c}
\toprule
 & \textbf{Cohort Type} & \textbf{Group} &   $N$ &    $n$ &  \textbf{Class} \\      
 & & & & & \textbf{Imbalance} \\
\midrule

\multirow{ 4 }{*}{\textbf{24 hours}} & \multirow{3}{*}{\textbf{Unsupervised}} &
  0 &  11862 &   404 &           0.0341 \\
   &    & 1 &   6434 &   107 &           0.0166 \\
   &    & 2 &  14390 &  1786 &           0.1241 \\ \cline{2-6}
   & \textbf{Global} & - &  32686 &  2297 &           0.0703 \\
\hline
\multirow{ 3 }{*}{\textbf{48 hours}} &\multirow{2}{*}{\textbf{Unsupervised}} & 0 &  13433 &   291 &           0.0217 \\
     &  & 1 &  16995 &  1436 &           0.0845 \\ \cline{2-6}
& \textbf{Global} & - &  30,428 &  1,727 &           0.0568 \\
\bottomrule
\end{tabular}
\end{centering}
}
\label{tab:cohorts}
\end{table}

\begin{table*}
\caption{24 Hour Mortality Prediction: Performance differences between multi-task and global models on specific cohorts. A multi-task model with pre-defined tasks based on careunits performs poorly, while the unsupervised multi-task model performs comparably on two out of three cohorts and better on one. Significant differences ($p < 0.01$) are shown in bold.}
\begin{threeparttable}
\centering
%\resizebox{0.7\textwidth}{!}{
\begin{tabular}{l | l | ll|ll|ll}
\toprule
          & & \multicolumn{2}{c|}{AUC} & \multicolumn{2}{c|}{PPV} & \multicolumn{2}{c}{Specificity} \\ \cline{3-8}
Cohort type & Cohort & Global & Multi-task & Global & Multi-task &  Global & Multi-task \\
\midrule 
\multirow{5}{*}{\textbf{Unsupervised}}  & 0 &  0.803 &    ${\bf 0.819}^{\dagger}$ &     0.083 &    ${\bf 0.103}^{\ddagger}$ &       0.732 &    ${\bf 0.786}^{\ddagger}$ \\
          & 1 &  0.811 &     ${\bf0.829}^{\dagger}$ &     0.120 &    ${\bf 0.126}^{\star}$ &       0.916 &     0.915 \\
          & 2 &  0.814 &    ${\bf 0.821}^{\ddagger}$ &     0.276 &   ${\bf  0.288}^{\ddagger}$ &       0.734 &    ${\bf 0.742}^{\ddagger}$ \\
          \cline{2-8}
          & Macro &  0.809 &    ${\bf  0.823}^{\dagger}$ &     0.159 &     ${\bf  0.172}^{\ddagger}$ &       0.794 &      ${\bf 0.814}^{\ddagger}$ \\ 	
          & Micro &    0.852  &    ${\bf 0.858}^{\ddagger}$  &      ${\bf 0.231}^{\diamond}$ &   0.228 &          ${\bf 0.817}^{\dagger}$ &    0.814 \\ \hline
\multirow{7}{*}{\textbf{Careunits}} & CCU &  ${\bf 0.862}^{\star}$  &     0.861 &     ${\bf 0.248}^{\ddagger}$ &     0.229 &       ${\bf 0.834}^{\ddagger}$ &     0.819 \\
          & CSRU &  0.849 &     ${\bf 0.867}^{\dagger}$ &     0.107 &     ${\bf 0.117 }^{\dagger}$ &       0.893 &    ${\bf  0.898}^{\dagger}$ \\
          & MICU &  0.814 &     ${\bf 0.832}^{\ddagger}$ &     0.261 &     ${\bf 0.262}^{\star}$ &       0.764 &    ${\bf  0.766}^{\star}$ \\
          & SICU &  0.839 &     ${\bf 0.855}^{\dagger}$ &     0.226 &     ${\bf 0.238}^{\dagger}$ &       0.781 &     ${\bf 0.796}^{\dagger}$ \\
          & TSICU &  0.846 &    ${\bf  0.869}^{\ddagger}$ &     0.183 &     ${\bf 0.192}^{\dagger}$ &       0.823 &     ${\bf 0.818}^{\diamond}$ \\
          \cline{2-8}
          & Macro &  0.842 &    ${\bf   0.857}^{\ddagger}$ &     0.205 &    ${\bf 0.208}^{\dagger}$ &       0.819 &   0.819 \\
          & Micro &    0.852 &     ${\bf  0.866}^{\ddagger}$ &     0.231 &     ${\bf  0.233}^{\diamond}$ &       0.817 &      ${\bf 0.821}^{\dagger}$ \\
\bottomrule
\end{tabular}
\centering
\begin{tablenotes}
      \small
      \item $\star$: 0.01 > $p$ > 0.001, $\diamond$: 0.001 > $p$ > 1e-5, $\dagger$: 1e-5 > $p$ > 1e-15, $\ddagger$: $p$ < 1e-15
\end{tablenotes}
%}
\label{breakdown_24hr_12gap}
\end{threeparttable}
\end{table*}

\subsection{Predicting Mortality at 24 Hours}
\subsubsection{Discovered Cohorts are Physiologically Distinct} Statistics about the discovered cohorts are shown in Table~\ref{tab:cohorts}, and Figure~\ref{phys_24_48} shows visualizations of the tasks learned using our methodology. 

Table~\ref{phys_24_48} shows that the three cohorts of patients discovered from the first 24 hours of data are different in terms of size and class imbalance. While two of the clusters are large, with over 10,000 patients each, the class imbalances in these two cohorts are dramatically different. Whereas Cohort 0 has an outcome incidence of 3\%, Cohort 1 has an outcome incidence of 12\%.

In addition, the centroids of the cohorts show physiological trends over the first 24 hours that differ in important ways (Figure~\ref{phys_24_48}). For example, clusters 0 and 2 both have elevated blood pressure in the first several hours of their stay. However, whereas cluster 0's blood pressure decreases over time, cluster 2's blood pressure stays elevated. We also observe that the heart rate in cluster 0 decreases over time, whereas cluster 1 and cluster 2 both have increasing heart rates. The differences between these cluster centroids indicate that our method of learning dense representations from the sparse physiological data for clustering discovers salient differences between patients. 

\subsubsection{Multi-task Models Outperform Global and Separate Models}
Our multi-task framework significantly improved performance over the global model in 
AUC, PPV, and specificity on each of the learned cohorts ($p$ < 0.01). In addition, it improved performance on aggregate metrics such as macro-AUC, PPV, and specificity, as well as micro-AUC. However, micro- PPV and micro-specificity were significantly worse. This is because micro-metrics are computed by setting a {\it single} threshold across all examples, regardless of the cohort they belong to. However, setting a single threshold ignores the large class imbalance differences between the cohorts. In contrast, the macro-measure, which considers a separate decision threshold based on 80\% sensitivity for each individual cohort, is significantly better when using the multi-task model compared to the global model. The performance increase from using the multi-task model indicates that we can improve both per-group and aggregate measures using this framework. 

More generally, we hope to highlight the importance of evaluating methods across subpopulations, since overall micro-measures can hide underperformance on specific subgroups. For example, the global model achieves an overall Micro AUC of 0.852, but it's AUC on cluster 0 was only 0.803. Without an evaluation broken down by groups, it would be hard to detect such performance disparities. 

We contrast our learned patient populations against expert knowledge driven cohorts, where patients are stratified by the first care unit they are admitted to. This cohort definition does not rely on the underlying physiological data. However, it is a reasonable attribute on which to split patients, given the differences across care units in patient conditions (see Figure~\ref{ccs_categories}). In addition, the rate of adverse events in these different units is highly variable, from less than two percent in the Cardiac Surgery Recovery unit to over 10\% in the Medical ICU. 

Grouping patients by first care unit and using these groups as tasks in an multi-task framework significantly improved performance over the global model. At this point in the patient's stay, first care unit is likely a meaningful indicator of differences between populations. However, while we have access to meaningful patient cohorts defined by first care unit, such distinct, labeled groups may not be available for a different clinical population. Our unsupervised method results in significant improvements, without requiring expert knowledge.

\begin{table*}[t!]
\caption{48 Hour Mortality Prediction: Performance differences between multi-task and global models on specific cohorts. A multi-task model with pre-defined tasks based on careunits performs poorly, while the unsupervised multi-task model performs comparably. Significant differences ($p < 0.01$) are shown in bold.}
\begin{threeparttable}
\centering
\begin{tabular}{l | l | cc|cc| cc}
\toprule
& & \multicolumn{2}{c|}{AUC} & \multicolumn{2}{c|}{PPV} & \multicolumn{2}{c}{Specificity} \\ \cline{3-8}
Cohort type & Cohort & Global & Multi-task & Global & Multi-task &  Global & Multi-task \\
\midrule
  \multirow{2}{*}{\textbf{Careunits}}   & Macro &  ${\bf 0.859}^{\ddagger}$  &     0.839 &    ${\bf 0.187}^{\ddagger}$ &     0.170 &      ${\bf 0.833}^{\ddagger}$ &     0.826 \\
       & Micro &    ${\bf 0.865}^{\ddagger}$ &     0.856 &     ${\bf   0.206}^{\dagger}$  &     0.198 &       ${\bf  0.833}^{\star}$ &     0.832 \\ \hline
\multirow{2}{*}{\textbf{Unsupervised}}          & Macro & 0.834 &     0.833 &     0.154 &     0.154 &     ${\bf   0.789}^{\ddagger}$ &     0.775 \\
          & Micro &   ${\bf  0.865}^{\dagger}$  &     0.861 &    0.206  &     0.191 &        ${\bf  0.833}^{\ddagger}$ &     0.812 \\
\bottomrule
\end{tabular}
\begin{tablenotes}
      \small
      \item $\star$: 0.01 > $p$ > 0.001, $\diamond$: 0.001 > $p$ > 1e-5, $\dagger$: 1e-5 > $p$ > 1e-15, $\ddagger$: $p < 1e-15$
\end{tablenotes}
\end{threeparttable}
\label{breakdown_48hrs_24gap}
\end{table*}

\subsection{Predicting Mortality at 48 Hours}

In contrast to the results from predicting mortality at 24 hours, our multi-task model with learned patient cohorts does not result in significant improvements compared to the global model when predicting mortality after 72 hours using 48 hours of data. One reason for this  may be the sparse nature of the physiological data. Because routine lab tests and other evaluations are frequently done in the first day of a patient's ICU stay, data presence drops off in the second day. Because of this, the data are heavily biased towards missing values; therefore, the autoencoder we use to construct dense representations of patient physiological state may also be biased.

The macro- and micro- performance metrics are shown in Table~\ref{breakdown_48hrs_24gap}. For our unsupervised method, macro AUC and PPV were not significantly different from the global model's performance, but the specificity was significantly worse. In addition, we again compared a multi-task model with learned cohorts against the expert-defined cohorts. In this case, while our method did not result in significant differences, the care units multi-task model performed significantly worse on all metrics compared to the global model. As a patient's stay in the ICU progresses, her characteristics may be less defined by the care unit she is admitted to compared to the interventions that are being administered. This highlights the need to use the underlying data to discover meaningful and distinct cohorts as tasks, and motivates further research on how to discover such cohorts in the presence of extreme sparsity (as in the 48-hour data).

\section{Conclusions \& Discussion}
\label{conclusions}

In this work, we show how machine learning models trained globally on heterogeneous populations can perform well in an overall sense while under-performing on specific, meaningful populations. We propose a two-step pipeline that 1) identifies distinct patient subpopulations, and 2) leverages these subpopulations in a multi-task framework to effectively share knowledge. 

We demonstrate that for 24-hour mortality prediction, our learned cohorts significantly improve over a single model learned on all of the data. In addition, we compare against an expert-knowledge driven method for identifying cohorts. We show that meaningful, distinct tasks can be learned in a data-driven way without pre-specifying cohorts for a particular outcome. We evaluate our models on the overall population, and on each separate cohort.

We highlight the need to evaluate performance across relevant cohorts. Much real data consists of heterogeneous populations, and reporting a single, overall evaluation metric can hide disparities in performance across groups. Accounting for these patient differences is important in model training, but also in model \textit{evaluation}.
%Our framework makes it straightforward in both training and evaluation to keep patient differences in mind. 

In addition, we believe that while unsupervised clustering of the physiological data representations led to improved results in the multi-task framework, learning clusters and representations that are guided by the specific outcome of interest could lead to useful outcome-specific cohorts. While unsupervised cohorts are generalizable across outcomes, representations and cohorts that are outcome-specific could lead to further improvements in predictive performance. For example, patient subpopulations that are distinct for predicting ventilator administration may look very different compared to patient subpopulations that are distinct for predicting length of stay or discharge status.

While the work we present is specific to the MIMIC-III dataset, we believe that the considerations we outline here are broadly applicable to clinical prediction tasks.  We hope the ideas we have discussed can help ensure that machine learning algorithms are not assumed to be one-size-fit-all, but rather that they work well for all groups involved.

\section{Acknowledgments}
This research was funded in part by the National Institutes of Health under award number P50DC015446 and by Wistron Corporation.
\bibliographystyle{ACM-Reference-Format}
\balance
\bibliography{bibliography}


\begin{thebibliography}{48}

%%% the Web address.

\ifx \showCODEN    \undefined \def \showCODEN     #1{\unskip}     \fi
\ifx \showDOI      \undefined \def \showDOI       #1{#1}\fi
\ifx \showISBNx    \undefined \def \showISBNx     #1{\unskip}     \fi
\ifx \showISBNxiii \undefined \def \showISBNxiii  #1{\unskip}     \fi
\ifx \showISSN     \undefined \def \showISSN      #1{\unskip}     \fi
\ifx \showLCCN     \undefined \def \showLCCN      #1{\unskip}     \fi
\ifx \shownote     \undefined \def \shownote      #1{#1}          \fi
\ifx \showarticletitle \undefined \def \showarticletitle #1{#1}   \fi
\ifx \showURL      \undefined \def \showURL       {\relax}        \fi
% invisible to TeX
\providecommand\bibfield[2]{#2}
\providecommand\bibinfo[2]{#2}
\providecommand\natexlab[1]{#1}
\providecommand\showeprint[2][]{arXiv:#2}

\bibitem[\protect\citeauthoryear{Alaa, Yoon, Hu, and van~der Schaar}{Alaa
  et~al\mbox{.}}{2016}]%
        {alaa2016personalized}
\bibfield{author}{\bibinfo{person}{Ahmed~M Alaa}, \bibinfo{person}{Jinsung
  Yoon}, \bibinfo{person}{Scott Hu}, {and} \bibinfo{person}{Mihaela van~der
  Schaar}.} \bibinfo{year}{2016}\natexlab{}.
\newblock \showarticletitle{Personalized risk scoring for critical care
  patients using mixtures of Gaussian Process Experts}.
\newblock \bibinfo{journal}{\emph{arXiv preprint arXiv:1605.00959}}
  (\bibinfo{year}{2016}).
\newblock


\bibitem[\protect\citeauthoryear{Bengio, Simard, and Frasconi}{Bengio
  et~al\mbox{.}}{1994}]%
        {bengio1994learning}
\bibfield{author}{\bibinfo{person}{Yoshua Bengio}, \bibinfo{person}{Patrice
  Simard}, {and} \bibinfo{person}{Paolo Frasconi}.}
  \bibinfo{year}{1994}\natexlab{}.
\newblock \showarticletitle{Learning long-term dependencies with gradient
  descent is difficult}.
\newblock \bibinfo{journal}{\emph{IEEE transactions on neural networks}}
  \bibinfo{volume}{5}, \bibinfo{number}{2} (\bibinfo{year}{1994}),
  \bibinfo{pages}{157--166}.
\newblock


\bibitem[\protect\citeauthoryear{Buolamwini and Gebru}{Buolamwini and
  Gebru}{2018}]%
        {pmlr-v81-buolamwini18a}
\bibfield{author}{\bibinfo{person}{Joy Buolamwini} {and}
  \bibinfo{person}{Timnit Gebru}.} \bibinfo{year}{2018}\natexlab{}.
\newblock \showarticletitle{Gender Shades: Intersectional Accuracy Disparities
  in Commercial Gender Classification}. In
  \bibinfo{booktitle}{\emph{Proceedings of the 1st Conference on Fairness,
  Accountability and Transparency}} \emph{(\bibinfo{series}{Proceedings of
  Machine Learning Research})}, \bibfield{editor}{\bibinfo{person}{Sorelle~A.
  Friedler} {and} \bibinfo{person}{Christo Wilson}} (Eds.),
  Vol.~\bibinfo{volume}{81}. \bibinfo{publisher}{PMLR}, \bibinfo{address}{New
  York, NY, USA}, \bibinfo{pages}{77--91}.
\newblock
\urldef\tempurl%
\url{http://proceedings.mlr.press/v81/buolamwini18a.html}
\showURL{%
\tempurl}


\bibitem[\protect\citeauthoryear{Caruana}{Caruana}{1993}]%
        {Caruana1993MultitaskLA}
\bibfield{author}{\bibinfo{person}{Rich Caruana}.}
  \bibinfo{year}{1993}\natexlab{}.
\newblock \showarticletitle{Multitask Learning: A Knowledge-Based Source of
  Inductive Bias}. In \bibinfo{booktitle}{\emph{ICML}}.
\newblock


\bibitem[\protect\citeauthoryear{Che, Kale, Li, Bahadori, and Liu}{Che
  et~al\mbox{.}}{2015}]%
        {che2015deep}
\bibfield{author}{\bibinfo{person}{Zhengping Che}, \bibinfo{person}{David
  Kale}, \bibinfo{person}{Wenzhe Li}, \bibinfo{person}{Mohammad~Taha Bahadori},
  {and} \bibinfo{person}{Yan Liu}.} \bibinfo{year}{2015}\natexlab{}.
\newblock \showarticletitle{Deep computational phenotyping}. In
  \bibinfo{booktitle}{\emph{Proceedings of the 21th ACM SIGKDD International
  Conference on Knowledge Discovery and Data Mining}}. ACM,
  \bibinfo{pages}{507--516}.
\newblock


\bibitem[\protect\citeauthoryear{Che, Purushotham, Cho, Sontag, and Liu}{Che
  et~al\mbox{.}}{2016}]%
        {che2016recurrent}
\bibfield{author}{\bibinfo{person}{Zhengping Che}, \bibinfo{person}{Sanjay
  Purushotham}, \bibinfo{person}{Kyunghyun Cho}, \bibinfo{person}{David
  Sontag}, {and} \bibinfo{person}{Yan Liu}.} \bibinfo{year}{2016}\natexlab{}.
\newblock \showarticletitle{Recurrent neural networks for multivariate time
  series with missing values}.
\newblock \bibinfo{journal}{\emph{arXiv preprint arXiv:1606.01865}}
  (\bibinfo{year}{2016}).
\newblock


\bibitem[\protect\citeauthoryear{Choi, Bahadori, Schuetz, Stewart, and
  Sun}{Choi et~al\mbox{.}}{2016}]%
        {choi2016doctor}
\bibfield{author}{\bibinfo{person}{Edward Choi}, \bibinfo{person}{Mohammad~Taha
  Bahadori}, \bibinfo{person}{Andy Schuetz}, \bibinfo{person}{Walter~F
  Stewart}, {and} \bibinfo{person}{Jimeng Sun}.}
  \bibinfo{year}{2016}\natexlab{}.
\newblock \showarticletitle{Doctor ai: Predicting clinical events via recurrent
  neural networks}. In \bibinfo{booktitle}{\emph{Machine Learning for
  Healthcare Conference}}. \bibinfo{pages}{301--318}.
\newblock


\bibitem[\protect\citeauthoryear{Chollet et~al\mbox{.}}{Chollet
  et~al\mbox{.}}{2015}]%
        {chollet2015keras}
\bibfield{author}{\bibinfo{person}{Fran\c{c}ois Chollet} {et~al\mbox{.}}}
  \bibinfo{year}{2015}\natexlab{}.
\newblock \bibinfo{title}{Keras}.
\newblock \bibinfo{howpublished}{\url{https://github.com/keras-team/keras}}.
  (\bibinfo{year}{2015}).
\newblock


\bibitem[\protect\citeauthoryear{Chorowski, Bahdanau, Serdyuk, Cho, and
  Bengio}{Chorowski et~al\mbox{.}}{2015}]%
        {chorowski2015attention}
\bibfield{author}{\bibinfo{person}{Jan~K Chorowski}, \bibinfo{person}{Dzmitry
  Bahdanau}, \bibinfo{person}{Dmitriy Serdyuk}, \bibinfo{person}{Kyunghyun
  Cho}, {and} \bibinfo{person}{Yoshua Bengio}.}
  \bibinfo{year}{2015}\natexlab{}.
\newblock \showarticletitle{Attention-based models for speech recognition}. In
  \bibinfo{booktitle}{\emph{Advances in Neural Information Processing
  Systems}}. \bibinfo{pages}{577--585}.
\newblock


\bibitem[\protect\citeauthoryear{Corbin, Richiardi, Vermeulen, Kromhout,
  Merletti, Peters, Simonato, Steenland, Pearce, and Maule}{Corbin
  et~al\mbox{.}}{2012}]%
        {corbin2012hierarchical}
\bibfield{author}{\bibinfo{person}{Marine Corbin}, \bibinfo{person}{Lorenzo
  Richiardi}, \bibinfo{person}{Roel Vermeulen}, \bibinfo{person}{Hans
  Kromhout}, \bibinfo{person}{Franco Merletti}, \bibinfo{person}{Susan Peters},
  \bibinfo{person}{Lorenzo Simonato}, \bibinfo{person}{Kyle Steenland},
  \bibinfo{person}{Neil Pearce}, {and} \bibinfo{person}{Milena Maule}.}
  \bibinfo{year}{2012}\natexlab{}.
\newblock \showarticletitle{Hierarchical regression for multiple comparisons in
  a case-control study of occupational risks for lung cancer}.
\newblock \bibinfo{journal}{\emph{PloS one}} \bibinfo{volume}{7},
  \bibinfo{number}{6} (\bibinfo{year}{2012}), \bibinfo{pages}{e38944}.
\newblock


\bibitem[\protect\citeauthoryear{Datta, Tschantz, and Datta}{Datta
  et~al\mbox{.}}{2015}]%
        {datta2015automated}
\bibfield{author}{\bibinfo{person}{Amit Datta}, \bibinfo{person}{Michael~Carl
  Tschantz}, {and} \bibinfo{person}{Anupam Datta}.}
  \bibinfo{year}{2015}\natexlab{}.
\newblock \showarticletitle{Automated experiments on ad privacy settings}.
\newblock \bibinfo{journal}{\emph{Proceedings on Privacy Enhancing
  Technologies}} \bibinfo{volume}{2015}, \bibinfo{number}{1}
  (\bibinfo{year}{2015}), \bibinfo{pages}{92--112}.
\newblock


\bibitem[\protect\citeauthoryear{D'Errigo, Tosti, Fusco, Perucci, and
  Seccareccia}{D'Errigo et~al\mbox{.}}{2007}]%
        {d2007use}
\bibfield{author}{\bibinfo{person}{Paola D'Errigo}, \bibinfo{person}{Maria~E
  Tosti}, \bibinfo{person}{Danilo Fusco}, \bibinfo{person}{Carlo~A Perucci},
  {and} \bibinfo{person}{Fulvia Seccareccia}.} \bibinfo{year}{2007}\natexlab{}.
\newblock \showarticletitle{Use of hierarchical models to evaluate performance
  of cardiac surgery centres in the Italian CABG outcome study}.
\newblock \bibinfo{journal}{\emph{BMC medical research methodology}}
  \bibinfo{volume}{7}, \bibinfo{number}{1} (\bibinfo{year}{2007}),
  \bibinfo{pages}{29}.
\newblock


\bibitem[\protect\citeauthoryear{Gehrmann, Dernoncourt, Li, Carlson, Wu, Welt,
  Foote~Jr, Moseley, Grant, Tyler, et~al\mbox{.}}{Gehrmann
  et~al\mbox{.}}{2017}]%
        {gehrmann2017comparing}
\bibfield{author}{\bibinfo{person}{Sebastian Gehrmann}, \bibinfo{person}{Franck
  Dernoncourt}, \bibinfo{person}{Yeran Li}, \bibinfo{person}{Eric~T Carlson},
  \bibinfo{person}{Joy~T Wu}, \bibinfo{person}{Jonathan Welt},
  \bibinfo{person}{John Foote~Jr}, \bibinfo{person}{Edward~T Moseley},
  \bibinfo{person}{David~W Grant}, \bibinfo{person}{Patrick~D Tyler},
  {et~al\mbox{.}}} \bibinfo{year}{2017}\natexlab{}.
\newblock \showarticletitle{Comparing Rule-Based and Deep Learning Models for
  Patient Phenotyping}.
\newblock \bibinfo{journal}{\emph{arXiv preprint arXiv:1703.08705}}
  (\bibinfo{year}{2017}).
\newblock


\bibitem[\protect\citeauthoryear{Ghassemi, Celi, and Stone}{Ghassemi
  et~al\mbox{.}}{2015}]%
        {ghassemi2015state}
\bibfield{author}{\bibinfo{person}{Marzyeh Ghassemi},
  \bibinfo{person}{Leo~Anthony Celi}, {and} \bibinfo{person}{David~J Stone}.}
  \bibinfo{year}{2015}\natexlab{}.
\newblock \showarticletitle{State of the art review: the data revolution in
  critical care}.
\newblock \bibinfo{journal}{\emph{Critical Care}} \bibinfo{volume}{19},
  \bibinfo{number}{1} (\bibinfo{year}{2015}), \bibinfo{pages}{118}.
\newblock


\bibitem[\protect\citeauthoryear{Ghassemi, Naumann, Doshi-Velez, Brimmer,
  Joshi, Rumshisky, and Szolovits}{Ghassemi et~al\mbox{.}}{2014}]%
        {ghassemi2014unfolding}
\bibfield{author}{\bibinfo{person}{Marzyeh Ghassemi}, \bibinfo{person}{Tristan
  Naumann}, \bibinfo{person}{Finale Doshi-Velez}, \bibinfo{person}{Nicole
  Brimmer}, \bibinfo{person}{Rohit Joshi}, \bibinfo{person}{Anna Rumshisky},
  {and} \bibinfo{person}{Peter Szolovits}.} \bibinfo{year}{2014}\natexlab{}.
\newblock \showarticletitle{Unfolding physiological state: Mortality modelling
  in intensive care units}. In \bibinfo{booktitle}{\emph{Proceedings of the
  20th ACM SIGKDD international conference on Knowledge discovery and data
  mining}}. ACM, \bibinfo{pages}{75--84}.
\newblock


\bibitem[\protect\citeauthoryear{Ghassemi, Wu, Hughes, Szolovits, and
  Doshi-Velez}{Ghassemi et~al\mbox{.}}{2017}]%
        {ghassemi2017predicting}
\bibfield{author}{\bibinfo{person}{Marzyeh Ghassemi}, \bibinfo{person}{Mike
  Wu}, \bibinfo{person}{Michael~C Hughes}, \bibinfo{person}{Peter Szolovits},
  {and} \bibinfo{person}{Finale Doshi-Velez}.} \bibinfo{year}{2017}\natexlab{}.
\newblock \showarticletitle{Predicting intervention onset in the ICU with
  switching state space models}.
\newblock \bibinfo{journal}{\emph{AMIA Summits on Translational Science
  Proceedings}} (\bibinfo{year}{2017}), \bibinfo{pages}{82}.
\newblock


\bibitem[\protect\citeauthoryear{Gong, Naumann, Szolovits, and Guttag}{Gong
  et~al\mbox{.}}{2017}]%
        {gong2017predicting}
\bibfield{author}{\bibinfo{person}{Jen~J Gong}, \bibinfo{person}{Tristan
  Naumann}, \bibinfo{person}{Peter Szolovits}, {and} \bibinfo{person}{John~V
  Guttag}.} \bibinfo{year}{2017}\natexlab{}.
\newblock \showarticletitle{Predicting Clinical Outcomes Across Changing
  Electronic Health Record Systems}. In \bibinfo{booktitle}{\emph{Proceedings
  of the 23rd ACM SIGKDD International Conference on Knowledge Discovery and
  Data Mining}}. ACM, \bibinfo{pages}{1497--1505}.
\newblock


\bibitem[\protect\citeauthoryear{Harutyunyan, Khachatrian, Kale, and
  Galstyan}{Harutyunyan et~al\mbox{.}}{2017}]%
        {harutyunyan2017multitask}
\bibfield{author}{\bibinfo{person}{Hrayr Harutyunyan}, \bibinfo{person}{Hrant
  Khachatrian}, \bibinfo{person}{David~C Kale}, {and} \bibinfo{person}{Aram
  Galstyan}.} \bibinfo{year}{2017}\natexlab{}.
\newblock \showarticletitle{Multitask Learning and Benchmarking with Clinical
  Time Series Data}.
\newblock \bibinfo{journal}{\emph{arXiv preprint arXiv:1703.07771}}
  (\bibinfo{year}{2017}).
\newblock


\bibitem[\protect\citeauthoryear{Hermann, Kocisky, Grefenstette, Espeholt, Kay,
  Suleyman, and Blunsom}{Hermann et~al\mbox{.}}{2015}]%
        {hermann2015teaching}
\bibfield{author}{\bibinfo{person}{Karl~Moritz Hermann}, \bibinfo{person}{Tomas
  Kocisky}, \bibinfo{person}{Edward Grefenstette}, \bibinfo{person}{Lasse
  Espeholt}, \bibinfo{person}{Will Kay}, \bibinfo{person}{Mustafa Suleyman},
  {and} \bibinfo{person}{Phil Blunsom}.} \bibinfo{year}{2015}\natexlab{}.
\newblock \showarticletitle{Teaching machines to read and comprehend}. In
  \bibinfo{booktitle}{\emph{Advances in Neural Information Processing
  Systems}}. \bibinfo{pages}{1693--1701}.
\newblock


\bibitem[\protect\citeauthoryear{Ho, Ghosh, Steinhubl, Stewart, Denny, Malin,
  and Sun}{Ho et~al\mbox{.}}{2014b}]%
        {ho2014limestone}
\bibfield{author}{\bibinfo{person}{Joyce~C Ho}, \bibinfo{person}{Joydeep
  Ghosh}, \bibinfo{person}{Steve~R Steinhubl}, \bibinfo{person}{Walter~F
  Stewart}, \bibinfo{person}{Joshua~C Denny}, \bibinfo{person}{Bradley~A
  Malin}, {and} \bibinfo{person}{Jimeng Sun}.}
  \bibinfo{year}{2014}\natexlab{b}.
\newblock \showarticletitle{Limestone: High-throughput candidate phenotype
  generation via tensor factorization}.
\newblock \bibinfo{journal}{\emph{Journal of biomedical informatics}}
  \bibinfo{volume}{52} (\bibinfo{year}{2014}), \bibinfo{pages}{199--211}.
\newblock


\bibitem[\protect\citeauthoryear{Ho, Ghosh, and Sun}{Ho et~al\mbox{.}}{2014a}]%
        {ho2014marble}
\bibfield{author}{\bibinfo{person}{Joyce~C Ho}, \bibinfo{person}{Joydeep
  Ghosh}, {and} \bibinfo{person}{Jimeng Sun}.}
  \bibinfo{year}{2014}\natexlab{a}.
\newblock \showarticletitle{Marble: high-throughput phenotyping from electronic
  health records via sparse nonnegative tensor factorization}. In
  \bibinfo{booktitle}{\emph{Proceedings of the 20th ACM SIGKDD international
  conference on Knowledge discovery and data mining}}. ACM,
  \bibinfo{pages}{115--124}.
\newblock


\bibitem[\protect\citeauthoryear{Hochreiter and Schmidhuber}{Hochreiter and
  Schmidhuber}{1997}]%
        {hochreiter1997long}
\bibfield{author}{\bibinfo{person}{Sepp Hochreiter} {and}
  \bibinfo{person}{J{\"u}rgen Schmidhuber}.} \bibinfo{year}{1997}\natexlab{}.
\newblock \showarticletitle{Long short-term memory}.
\newblock \bibinfo{journal}{\emph{Neural computation}} \bibinfo{volume}{9},
  \bibinfo{number}{8} (\bibinfo{year}{1997}), \bibinfo{pages}{1735--1780}.
\newblock


\bibitem[\protect\citeauthoryear{Johnson, Pollard, Shen, Li-wei, Feng,
  Ghassemi, Moody, Szolovits, Celi, and Mark}{Johnson et~al\mbox{.}}{2016}]%
        {johnson2016mimic}
\bibfield{author}{\bibinfo{person}{Alistair~EW Johnson}, \bibinfo{person}{Tom~J
  Pollard}, \bibinfo{person}{Lu Shen}, \bibinfo{person}{H~Lehman Li-wei},
  \bibinfo{person}{Mengling Feng}, \bibinfo{person}{Mohammad Ghassemi},
  \bibinfo{person}{Benjamin Moody}, \bibinfo{person}{Peter Szolovits},
  \bibinfo{person}{Leo~Anthony Celi}, {and} \bibinfo{person}{Roger~G Mark}.}
  \bibinfo{year}{2016}\natexlab{}.
\newblock \showarticletitle{MIMIC-III, a freely accessible critical care
  database}.
\newblock \bibinfo{journal}{\emph{Scientific data}}  \bibinfo{volume}{3}
  (\bibinfo{year}{2016}), \bibinfo{pages}{160035}.
\newblock


\bibitem[\protect\citeauthoryear{Le~Gall, Lemeshow, and Saulnier}{Le~Gall
  et~al\mbox{.}}{1993}]%
        {le1993new}
\bibfield{author}{\bibinfo{person}{Jean-Roger Le~Gall},
  \bibinfo{person}{Stanley Lemeshow}, {and} \bibinfo{person}{Fabienne
  Saulnier}.} \bibinfo{year}{1993}\natexlab{}.
\newblock \showarticletitle{A new simplified acute physiology score (SAPS II)
  based on a European/North American multicenter study}.
\newblock \bibinfo{journal}{\emph{Jama}} \bibinfo{volume}{270},
  \bibinfo{number}{24} (\bibinfo{year}{1993}), \bibinfo{pages}{2957--2963}.
\newblock


\bibitem[\protect\citeauthoryear{Lipton, Kale, Elkan, and Wetzell}{Lipton
  et~al\mbox{.}}{2015}]%
        {lipton2015learning}
\bibfield{author}{\bibinfo{person}{Zachary~C Lipton}, \bibinfo{person}{David~C
  Kale}, \bibinfo{person}{Charles Elkan}, {and} \bibinfo{person}{Randall
  Wetzell}.} \bibinfo{year}{2015}\natexlab{}.
\newblock \showarticletitle{Learning to diagnose with LSTM recurrent neural
  networks}.
\newblock \bibinfo{journal}{\emph{arXiv preprint arXiv:1511.03677}}
  (\bibinfo{year}{2015}).
\newblock


\bibitem[\protect\citeauthoryear{Manning, Raghavan, and Schütze}{Manning
  et~al\mbox{.}}{2008}]%
        {macro_auc}
\bibfield{author}{\bibinfo{person}{Christopher Manning},
  \bibinfo{person}{Prabhakar Raghavan}, {and} \bibinfo{person}{Hinrich
  Schütze}.} \bibinfo{year}{2008}\natexlab{}.
\newblock \bibinfo{booktitle}{\emph{Introduction to Information Retrieval}}.
\newblock \bibinfo{publisher}{Cambridge University Press}.
\newblock


\bibitem[\protect\citeauthoryear{Miotto, Li, Kidd, and Dudley}{Miotto
  et~al\mbox{.}}{2016}]%
        {miotto2016deep}
\bibfield{author}{\bibinfo{person}{Riccardo Miotto}, \bibinfo{person}{Li Li},
  \bibinfo{person}{Brian~A Kidd}, {and} \bibinfo{person}{Joel~T Dudley}.}
  \bibinfo{year}{2016}\natexlab{}.
\newblock \showarticletitle{Deep patient: an unsupervised representation to
  predict the future of patients from the electronic health records}.
\newblock \bibinfo{journal}{\emph{Scientific reports}}  \bibinfo{volume}{6}
  (\bibinfo{year}{2016}), \bibinfo{pages}{26094}.
\newblock


\bibitem[\protect\citeauthoryear{Ngufor, Upadhyaya, Murphree, Kor, and
  Pathak}{Ngufor et~al\mbox{.}}{2015a}]%
        {7344836}
\bibfield{author}{\bibinfo{person}{C. Ngufor}, \bibinfo{person}{S. Upadhyaya},
  \bibinfo{person}{D. Murphree}, \bibinfo{person}{D. Kor}, {and}
  \bibinfo{person}{J. Pathak}.} \bibinfo{year}{2015}\natexlab{a}.
\newblock \showarticletitle{Multi-task learning with selective cross-task
  transfer for predicting bleeding and other important patient outcomes}. In
  \bibinfo{booktitle}{\emph{2015 IEEE International Conference on Data Science
  and Advanced Analytics (DSAA)}}. \bibinfo{pages}{1--8}.
\newblock
\urldef\tempurl%
\url{https://doi.org/10.1109/DSAA.2015.7344836}
\showDOI{\tempurl}


\bibitem[\protect\citeauthoryear{Ngufor, Upadhyaya, Murphree, Madde, Kor, and
  Pathak}{Ngufor et~al\mbox{.}}{2015b}]%
        {ngufor2015heterogeneous}
\bibfield{author}{\bibinfo{person}{Che Ngufor}, \bibinfo{person}{Sudhindra
  Upadhyaya}, \bibinfo{person}{Dennis Murphree}, \bibinfo{person}{Nageswar
  Madde}, \bibinfo{person}{Daryl Kor}, {and} \bibinfo{person}{Jyotishman
  Pathak}.} \bibinfo{year}{2015}\natexlab{b}.
\newblock \showarticletitle{A heterogeneous multi-task learning for predicting
  RBC transfusion and perioperative outcomes}. In
  \bibinfo{booktitle}{\emph{Conference on Artificial Intelligence in Medicine
  in Europe}}. Springer, \bibinfo{pages}{287--297}.
\newblock


\bibitem[\protect\citeauthoryear{Nori, Kashima, Yamashita, Kunisawa, and
  Imanaka}{Nori et~al\mbox{.}}{2017}]%
        {nori2017learning}
\bibfield{author}{\bibinfo{person}{Nozomi Nori}, \bibinfo{person}{Hisashi
  Kashima}, \bibinfo{person}{Kazuto Yamashita}, \bibinfo{person}{Susumu
  Kunisawa}, {and} \bibinfo{person}{Yuichi Imanaka}.}
  \bibinfo{year}{2017}\natexlab{}.
\newblock \showarticletitle{Learning Implicit Tasks for Patient-Specific Risk
  Modeling in ICU.}. In \bibinfo{booktitle}{\emph{AAAI}}.
  \bibinfo{pages}{1481--1487}.
\newblock


\bibitem[\protect\citeauthoryear{Obermeyer and Emanuel}{Obermeyer and
  Emanuel}{2016}]%
        {obermeyer2016predicting}
\bibfield{author}{\bibinfo{person}{Ziad Obermeyer} {and}
  \bibinfo{person}{Ezekiel~J Emanuel}.} \bibinfo{year}{2016}\natexlab{}.
\newblock \showarticletitle{{Predicting the Future---Big Data, Machine
  Learning, and Clinical Medicine}}.
\newblock \bibinfo{journal}{\emph{The New England journal of medicine}}
  \bibinfo{volume}{375}, \bibinfo{number}{13} (\bibinfo{year}{2016}),
  \bibinfo{pages}{1216}.
\newblock


\bibitem[\protect\citeauthoryear{Pedregosa, Varoquaux, Gramfort, Michel,
  Thirion, Grisel, Blondel, Prettenhofer, Weiss, Dubourg, Vanderplas, Passos,
  Cournapeau, Brucher, Perrot, and Duchesnay}{Pedregosa et~al\mbox{.}}{2011}]%
        {scikit-learn}
\bibfield{author}{\bibinfo{person}{F. Pedregosa}, \bibinfo{person}{G.
  Varoquaux}, \bibinfo{person}{A. Gramfort}, \bibinfo{person}{V. Michel},
  \bibinfo{person}{B. Thirion}, \bibinfo{person}{O. Grisel},
  \bibinfo{person}{M. Blondel}, \bibinfo{person}{P. Prettenhofer},
  \bibinfo{person}{R. Weiss}, \bibinfo{person}{V. Dubourg}, \bibinfo{person}{J.
  Vanderplas}, \bibinfo{person}{A. Passos}, \bibinfo{person}{D. Cournapeau},
  \bibinfo{person}{M. Brucher}, \bibinfo{person}{M. Perrot}, {and}
  \bibinfo{person}{E. Duchesnay}.} \bibinfo{year}{2011}\natexlab{}.
\newblock \showarticletitle{Scikit-learn: Machine Learning in {P}ython}.
\newblock \bibinfo{journal}{\emph{Journal of Machine Learning Research}}
  \bibinfo{volume}{12} (\bibinfo{year}{2011}), \bibinfo{pages}{2825--2830}.
\newblock


\bibitem[\protect\citeauthoryear{Pivovarov, Perotte, Grave, Angiolillo,
  Wiggins, and Elhadad}{Pivovarov et~al\mbox{.}}{2015}]%
        {pivovarov2015learning}
\bibfield{author}{\bibinfo{person}{Rimma Pivovarov}, \bibinfo{person}{Adler~J
  Perotte}, \bibinfo{person}{Edouard Grave}, \bibinfo{person}{John Angiolillo},
  \bibinfo{person}{Chris~H Wiggins}, {and} \bibinfo{person}{No{\'e}mie
  Elhadad}.} \bibinfo{year}{2015}\natexlab{}.
\newblock \showarticletitle{Learning probabilistic phenotypes from
  heterogeneous EHR data}.
\newblock \bibinfo{journal}{\emph{Journal of biomedical informatics}}
  \bibinfo{volume}{58} (\bibinfo{year}{2015}), \bibinfo{pages}{156--165}.
\newblock


\bibitem[\protect\citeauthoryear{Razavian, Marcus, and Sontag}{Razavian
  et~al\mbox{.}}{2016}]%
        {razavian2016multi}
\bibfield{author}{\bibinfo{person}{Narges Razavian}, \bibinfo{person}{Jake
  Marcus}, {and} \bibinfo{person}{David Sontag}.}
  \bibinfo{year}{2016}\natexlab{}.
\newblock \showarticletitle{Multi-task prediction of disease onsets from
  longitudinal laboratory tests}. In \bibinfo{booktitle}{\emph{Machine Learning
  for Healthcare Conference}}. \bibinfo{pages}{73--100}.
\newblock


\bibitem[\protect\citeauthoryear{Ruder}{Ruder}{2017}]%
        {DBLP:journals/corr/Ruder17a}
\bibfield{author}{\bibinfo{person}{Sebastian Ruder}.}
  \bibinfo{year}{2017}\natexlab{}.
\newblock \showarticletitle{An Overview of Multi-Task Learning in Deep Neural
  Networks}.
\newblock \bibinfo{journal}{\emph{CoRR}}  \bibinfo{volume}{abs/1706.05098}
  (\bibinfo{year}{2017}).
\newblock
\showeprint[arxiv]{1706.05098}
\urldef\tempurl%
\url{http://arxiv.org/abs/1706.05098}
\showURL{%
\tempurl}


\bibitem[\protect\citeauthoryear{Shaddox, Ryan, Schuemie, Madigan, and
  Suchard}{Shaddox et~al\mbox{.}}{2016}]%
        {shaddox2016hierarchical}
\bibfield{author}{\bibinfo{person}{Trevor~R Shaddox},
  \bibinfo{person}{Patrick~B Ryan}, \bibinfo{person}{Martijn~J Schuemie},
  \bibinfo{person}{David Madigan}, {and} \bibinfo{person}{Marc~A Suchard}.}
  \bibinfo{year}{2016}\natexlab{}.
\newblock \showarticletitle{Hierarchical models for multiple, rare outcomes
  using massive observational healthcare databases}.
\newblock \bibinfo{journal}{\emph{Statistical Analysis and Data Mining: The ASA
  Data Science Journal}} \bibinfo{volume}{9}, \bibinfo{number}{4}
  (\bibinfo{year}{2016}), \bibinfo{pages}{260--268}.
\newblock


\bibitem[\protect\citeauthoryear{Shankar, Halpern, Breck, Atwood, Wilson, and
  Sculley}{Shankar et~al\mbox{.}}{2017}]%
        {46553}
\bibfield{author}{\bibinfo{person}{Shreya Shankar}, \bibinfo{person}{Yoni
  Halpern}, \bibinfo{person}{Eric Breck}, \bibinfo{person}{James Atwood},
  \bibinfo{person}{Jimbo Wilson}, {and} \bibinfo{person}{D. Sculley}.}
  \bibinfo{year}{2017}\natexlab{}.
\newblock \showarticletitle{No Classification without Representation: Assessing
  Geodiversity Issues in Open Data Sets for the Developing World}. In
  \bibinfo{booktitle}{\emph{NIPS 2017 workshop: Machine Learning for the
  Developing World}}.
\newblock


\bibitem[\protect\citeauthoryear{Sinuff, Adhikari, Cook, Sch{\"u}nemann,
  Griffith, Rocker, and Walter}{Sinuff et~al\mbox{.}}{2006}]%
        {sinuff2006mortality}
\bibfield{author}{\bibinfo{person}{Tasnim Sinuff}, \bibinfo{person}{Neill~KJ
  Adhikari}, \bibinfo{person}{Deborah~J Cook}, \bibinfo{person}{Holger~J
  Sch{\"u}nemann}, \bibinfo{person}{Lauren~E Griffith}, \bibinfo{person}{Graeme
  Rocker}, {and} \bibinfo{person}{Stephen~D Walter}.}
  \bibinfo{year}{2006}\natexlab{}.
\newblock \showarticletitle{Mortality predictions in the intensive care unit:
  comparing physicians with scoring systems}.
\newblock \bibinfo{journal}{\emph{Critical care medicine}}
  \bibinfo{volume}{34}, \bibinfo{number}{3} (\bibinfo{year}{2006}),
  \bibinfo{pages}{878--885}.
\newblock


\bibitem[\protect\citeauthoryear{Suresh, Hunt, Johnson, Celi, Szolovits, and
  Ghassemi}{Suresh et~al\mbox{.}}{2017a}]%
        {suresh2017clinical}
\bibfield{author}{\bibinfo{person}{Harini Suresh}, \bibinfo{person}{Nathan
  Hunt}, \bibinfo{person}{Alistair Johnson}, \bibinfo{person}{Leo~Anthony
  Celi}, \bibinfo{person}{Peter Szolovits}, {and} \bibinfo{person}{Marzyeh
  Ghassemi}.} \bibinfo{year}{2017}\natexlab{a}.
\newblock \showarticletitle{Clinical Intervention Prediction and Understanding
  using Deep Networks}.
\newblock \bibinfo{journal}{\emph{Machine Learning for Health}}
  (\bibinfo{year}{2017}).
\newblock


\bibitem[\protect\citeauthoryear{Suresh, Szolovits, and Ghassemi}{Suresh
  et~al\mbox{.}}{2017b}]%
        {suresh2017use}
\bibfield{author}{\bibinfo{person}{Harini Suresh}, \bibinfo{person}{Peter
  Szolovits}, {and} \bibinfo{person}{Marzyeh Ghassemi}.}
  \bibinfo{year}{2017}\natexlab{b}.
\newblock \showarticletitle{The use of autoencoders for discovering patient
  phenotypes}.
\newblock \bibinfo{journal}{\emph{arXiv preprint arXiv:1703.07004}}
  (\bibinfo{year}{2017}).
\newblock


\bibitem[\protect\citeauthoryear{Wang, Wang, Hu, and Sorrentino}{Wang
  et~al\mbox{.}}{2014}]%
        {wang2014exploring}
\bibfield{author}{\bibinfo{person}{Xiang Wang}, \bibinfo{person}{Fei Wang},
  \bibinfo{person}{Jianying Hu}, {and} \bibinfo{person}{Robert Sorrentino}.}
  \bibinfo{year}{2014}\natexlab{}.
\newblock \showarticletitle{Exploring joint disease risk prediction}. In
  \bibinfo{booktitle}{\emph{AMIA Annual Symposium Proceedings}},
  Vol.~\bibinfo{volume}{2014}. American Medical Informatics Association,
  \bibinfo{pages}{1180}.
\newblock


\bibitem[\protect\citeauthoryear{Wiens, Guttag, and Horvitz}{Wiens
  et~al\mbox{.}}{2016}]%
        {wiens2016patient}
\bibfield{author}{\bibinfo{person}{Jenna Wiens}, \bibinfo{person}{John Guttag},
  {and} \bibinfo{person}{Eric Horvitz}.} \bibinfo{year}{2016}\natexlab{}.
\newblock \showarticletitle{Patient risk stratification with time-varying
  parameters: a multitask learning approach}.
\newblock \bibinfo{journal}{\emph{The Journal of Machine Learning Research}}
  \bibinfo{volume}{17}, \bibinfo{number}{1} (\bibinfo{year}{2016}),
  \bibinfo{pages}{2797--2819}.
\newblock


\bibitem[\protect\citeauthoryear{Wilcoxon}{Wilcoxon}{1945}]%
        {wilcoxon1945individual}
\bibfield{author}{\bibinfo{person}{Frank Wilcoxon}.}
  \bibinfo{year}{1945}\natexlab{}.
\newblock \showarticletitle{Individual comparisons by ranking methods}.
\newblock \bibinfo{journal}{\emph{Biometrics}} \bibinfo{volume}{1},
  \bibinfo{number}{6} (\bibinfo{year}{1945}), \bibinfo{pages}{80--83}.
\newblock


\bibitem[\protect\citeauthoryear{Wu, Roy, and Stewart}{Wu
  et~al\mbox{.}}{2010}]%
        {wu2010prediction}
\bibfield{author}{\bibinfo{person}{Jionglin Wu}, \bibinfo{person}{Jason Roy},
  {and} \bibinfo{person}{Walter~F Stewart}.} \bibinfo{year}{2010}\natexlab{}.
\newblock \showarticletitle{Prediction modeling using EHR data: challenges,
  strategies, and a comparison of machine learning approaches}.
\newblock \bibinfo{journal}{\emph{Medical care}} \bibinfo{volume}{48},
  \bibinfo{number}{6} (\bibinfo{year}{2010}), \bibinfo{pages}{S106--S113}.
\newblock


\bibitem[\protect\citeauthoryear{Wu, Ghassemi, Feng, Celi, Szolovits, and
  Doshi-Velez}{Wu et~al\mbox{.}}{2017}]%
        {wu2017understanding}
\bibfield{author}{\bibinfo{person}{Mike Wu}, \bibinfo{person}{Marzyeh
  Ghassemi}, \bibinfo{person}{Mengling Feng}, \bibinfo{person}{Leo~A Celi},
  \bibinfo{person}{Peter Szolovits}, {and} \bibinfo{person}{Finale
  Doshi-Velez}.} \bibinfo{year}{2017}\natexlab{}.
\newblock \showarticletitle{Understanding vasopressor intervention and weaning:
  Risk prediction in a public heterogeneous clinical time series database}.
\newblock \bibinfo{journal}{\emph{Journal of the American Medical Informatics
  Association}} \bibinfo{volume}{24}, \bibinfo{number}{3}
  (\bibinfo{year}{2017}), \bibinfo{pages}{488--495}.
\newblock


\bibitem[\protect\citeauthoryear{Xu, Zhou, and Tan}{Xu et~al\mbox{.}}{2015b}]%
        {xu2015formula}
\bibfield{author}{\bibinfo{person}{Jianpeng Xu}, \bibinfo{person}{Jiayu Zhou},
  {and} \bibinfo{person}{Pang-Ning Tan}.} \bibinfo{year}{2015}\natexlab{b}.
\newblock \showarticletitle{Formula: F act OR ized MU lti-task L e A rning for
  task discovery in personalized medical models}. In
  \bibinfo{booktitle}{\emph{Proceedings of the 2015 SIAM International
  Conference on Data Mining}}. SIAM, \bibinfo{pages}{496--504}.
\newblock


\bibitem[\protect\citeauthoryear{Xu, Ba, Kiros, Cho, Courville, Salakhutdinov,
  Zemel, and Bengio}{Xu et~al\mbox{.}}{2015a}]%
        {xu2015show}
\bibfield{author}{\bibinfo{person}{Kelvin Xu}, \bibinfo{person}{Jimmy Ba},
  \bibinfo{person}{Ryan Kiros}, \bibinfo{person}{Kyunghyun Cho},
  \bibinfo{person}{Aaron~C Courville}, \bibinfo{person}{Ruslan Salakhutdinov},
  \bibinfo{person}{Richard~S Zemel}, {and} \bibinfo{person}{Yoshua Bengio}.}
  \bibinfo{year}{2015}\natexlab{a}.
\newblock \showarticletitle{Show, Attend and Tell: Neural Image Caption
  Generation with Visual Attention.}. In \bibinfo{booktitle}{\emph{ICML}},
  Vol.~\bibinfo{volume}{14}. \bibinfo{pages}{77--81}.
\newblock


\bibitem[\protect\citeauthoryear{Zhang and Yang}{Zhang and Yang}{2017}]%
        {zhang2017survey}
\bibfield{author}{\bibinfo{person}{Yu Zhang} {and} \bibinfo{person}{Qiang
  Yang}.} \bibinfo{year}{2017}\natexlab{}.
\newblock \showarticletitle{A survey on multi-task learning}.
\newblock \bibinfo{journal}{\emph{arXiv preprint arXiv:1707.08114}}
  (\bibinfo{year}{2017}).
\newblock


\end{thebibliography}
\end{document}